\documentclass[10pt, a4paper]{article}
\usepackage{booktabs}
\usepackage{todonotes}
\usepackage{amsmath}
\usepackage{amssymb}
\usepackage{comment}
\usepackage{multirow}

\usepackage{xurl}
\usepackage[hidelinks]{hyperref}
\usepackage{cleveref}

\usepackage[final]{lrec2026} 

\title{Environmental, Social and Governance Sentiment Analysis on Slovene News: A Novel Dataset and Models} 

\name{%
\parbox{\textwidth}{\centering\bfseries
Paula Dodig$^{1}$ \quad
Boshko Koloski$^{2}$ \quad
Katarina Sitar Šuštar$^{3}$ \\
Senja Pollak$^{2}$ \quad
Matthew Purver$^{2,4}$
}%
}

\address{%
\parbox{\textwidth}{\centering\small
$^{1}$ Eindhoven University of Technology, Eindhoven \\
$^{2}$ Jožef Stefan Institute and Postgraduate School, Ljubljana \\
$^{3}$ Faculty of Economics, University of Ljubljana \\
$^{2}$ Queen Mary University of London \\
p.dodig@student.tue.nl, \{boshko.koloski, senja.pollak\}@ijs.si, \\ katarina.sitar@ef.uni-lj.si, m.purver@qmul.ac.uk
}%
}
\abstract{
Environmental, Social, and Governance (ESG) considerations are increasingly integral to assessing corporate performance, reputation, and long-term sustainability. Yet, reliable ESG ratings remain limited for smaller companies and emerging markets. 
We introduce the first publicly available Slovene ESG sentiment dataset and a suite of models for automatic ESG sentiment detection. The dataset, derived from the MaCoCu Slovene news collection, combines large language model (LLM)-assisted filtering with human annotation of company-related ESG content. We 
evaluate the performance of monolingual (SloBERTa) and multilingual (XLM-R) models, embedding-based classifiers (TabPFN), hierarchical ensemble architectures, and large language models. Results show that LLMs achieve the strongest performance on Environmental (Gemma3-27B, F1-macro: 0.61) and Social aspects (gpt-oss 20B, F1-macro: 0.45), while fine-tuned SloBERTa is the best model on Governance classification (F1-macro: 0.54). We then show in a small case study how the best-preforming classifier (gpt-oss) can be applied to investigate ESG aspects for selected companies across a long time frame.  
 \\ \newline  }

\begin{document}

\maketitleabstract
\pagenumbering{gobble}

\section{Introduction}
Environmental, Social, and Governance (ESG) considerations have become essential in the evaluation of corporate performance and investment potential \cite{chen2023environmental}. Increased awareness of corporate sustainability has led to the integration of ESG metrics into financial and public evaluations of businesses. Despite this momentum, a significant number of smaller publicly listed companies lack formal ESG ratings, making it difficult for ESG-focused retail investors to assess their sustainability performance \citep{bazrafshan2023role}. 
Moreover, existing ESG ratings are typically static, updated infrequently, and therefore unable to capture short-term shifts in public or media perception. They also tend to aggregate information from limited, often homogeneous sources, which obscures variation in how companies are portrayed across different news outlets and domains. Consequently, traditional ESG ratings fail to provide a dynamic or diversified view of corporate reputation as it evolves in real time. This gap is particularly pronounced in less-resourced linguistic contexts, where limited data availability and language-specific barriers further hinder analysis. Our research addresses this challenge by developing an automated, sentiment-based framework leveraging large language models (LLMs) to evaluate ESG-related content in news articles, with a specific focus on Slovene — a less-resourced Slavic language.

The research addresses the following questions. First, can we develop an automated, sentiment-based framework for ESG aspects in Slovenian textual data, more specifically in Slovenian news? Second, can we use this framework to track ESG-related perception of companies through associated news text? 


The main contributions of this paper are as follows. First, we present the first sentiment-annotated dataset from Slovenian news media on the aspects of Environment, Social and Governance considerations (\textbf{SloESG-News 1.0}). The development of this dataset, is based on the MaCoCu Slovene News dataset \citep{non-etal-2022-macocu} 
and uses the IPTC news codes and large language models (LLMs) for selection of articles for annotation. The gold standard annotation is provided by human annotators, resulting in a new publicly available resource for Slovenian.
Next, the dataset is used for training \textbf{ESG sentiment models} for Slovenian and evaluating their performance.
By extensive set of experiments using fine-tuned monolingual and multilingual transformers, LLMs, embedding-based classifier and hierarchical ensembles, we provide a replicable methodology and select the best models for each aspect. Third, we use the selected models in a \textbf{case study} with an expert from the field of economics. We apply the ESG model on a corpus of Slovene news for selected companies across 15 years, and show the change in E, S and G sentiment across time. The contextualisation and interpretation of results shows the potential of our method for interdisciplinary research and further more detailed qualitative case studies.

\section{Related Work}
The increasing relevance of ESG topics has driven the development of computational methods for understanding sustainability discourse in text. Prior research on ESG-related text analysis has focused on company reports and financial disclosures, leveraging supervised machine learning to assess sentiment and topic relevance \citep{nassirtoussi2015text}. However, such approaches are often limited by the availability of labeled data and by their focus on English and other high-resource languages.

Recent advances in transformer-based language models, such as BERT \citep{devlin2019bert} and RoBERTa \citep{liu2019roberta}, have enabled more nuanced sentiment and topic classification across domains, including finance and social responsibility \citep{araci2019finbert}. 


The application of NLP to ESG analysis has gained significant momentum, with transformer-based models proving particularly effective for processing corporate disclosures and news articles at scale~\citep{schimanski2024bridging}. Domain-specific models like FinBERT-ESG and ESGBERT have been developed through fine-tuning on ESG-specific corpora, achieving strong performance across environmental, social, and governance classification tasks~\citep{araci2019finbert,mehra2022esgbert}. Recent work has explored knowledge-enhanced approaches, with~\citet{koloski-etal-2022-knowledge} proposing representations that combine knowledge graphs and taxonomies with document embeddings for sustainability detection, while~\citet{angioni2024exploring} employed knowledge graphs to track ESG discourse evolution in news articles. BERT-based sentiment analysis has demonstrated predictive power for market reactions, with positive ESG news correlating with average abnormal returns of 0.31\% and negative news with -0.75\%~\citep{dorfleitner2024esg}. The FinNLP workshop series has hosted multilingual ESG research through shared tasks on ESG issue identification across multiple languages~\citep{tseng2023dynamicesg}, while recent studies have integrated ESG sentiment with technical indicators for financial forecasting~\citep{lee2024deep}. However, most work focuses on English and high-resource languages, with limited research on low-resource contexts like Slovene.


Recenlty, LLMs have been explored as tools for dataset curation and pseudo-labeling. The teacher–student framework for topic classification \citep{kuzman2025teacherstudent} has been shown to produce reliable results with minimal manual supervision, especially for under-resourced languages. Building on these insights, our study applies LLM-assisted filtering and human validation to create the first Slovene ESG sentiment dataset.

\section{SloESG-News 1.0 dataset}


To create an appropriate dataset, we extracted articles from the MaCoCu Slovenian dataset \citep{non-etal-2022-macocu}, filtering for a curated list of Slovenian companies, defined by an expert in ESG focusing on companies where at least one of the three aspects E, S or G is strongly present. The data was preprocessed to extract a subset of news articles where company names and ESG-related terminology co-occurred, by applying the Slavic-XLMR named entity recognition model~\cite{ivacic-etal-2023-analysis}, together with the IPTC media topic classifier from the CLASSLA repository~\citep{kuzman2025teacherstudent}. Manual annotation was then conducted in collaboration with economics students from the University of Ljubljana, who were trained to tag sentiment (positive, neutral, negative, or irrelevant) separately for each ESG aspect (Environmental, Social, Governance). The sentiment label was assessed specifically from the point of view of the company mentioned in the text (thus necessitating the inclusion of the ``irrelevant'' category). Initially, 24 annotators were considered, each given a set of 40 articles, 30 unique for individual annotation, and 10 shared articles jointly annotated by everyone. Along with student annotators, an expert annotation was used to identify outliers. After a student-expert pairwise agreement was calculated, 6 of the student annotators were discarded from the dataset due to a low agreement level, signifying faulty annotations and outlier behavior. The final dataset consists of 550 unique articles, where 10 articles were annotated by 19 annotators, while 540 by a single annotator.

\begin{figure}[h!]
    \centering
    \includegraphics[width=0.5\textwidth]{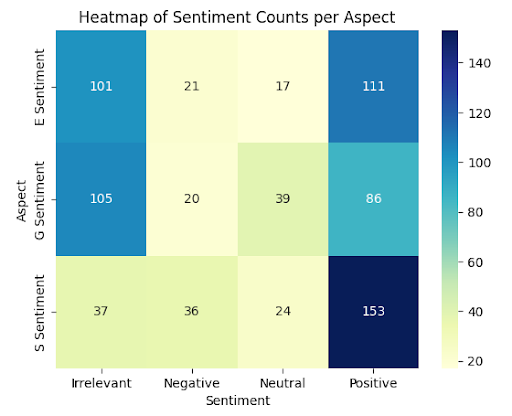}
    \caption{Student annotation results}
    \label{fig:heatmap}
\end{figure}

Ten articles were annotated by all annotators, allowing us to calculate inter-annotator agreement using  Fleiss' \textit{kappa} metric for multiple annotators \cite{fleiss1973equivalence}. This highlighted the inherent complexity of ESG sentiment interpretation: while the E category showed very strong agreement (close to 0.8), S agreement was in the moderate range (0.4) and we saw only low agreement on the G category (0.2). A heatmap of sentiment counts can be seen in Gigure \ref{fig:heatmap}. 

The dataset is split into training and test parts (see Table~\ref{tab:dataset_distribution}) and will be made available on CLARIN upon acceptance.


\begin{table}[t]
\centering
\caption{Dataset distribution.}
\label{tab:dataset_distribution}
\small
\begin{tabular}{llrrrr}
\toprule
\textbf{Split} & \textbf{Aspect} & \textbf{Irrel.} & \textbf{Neg.} & \textbf{Neut.} & \textbf{Pos.} \\
\midrule
\multirow{3}{*}{Train (440)} 
& E & 288 & 41 & 39 & 72 \\
& S & 144 & 48 & 69 & 179 \\
& G & 193 & 78 & 86 & 83 \\
\midrule
\multirow{3}{*}{Test (110)} 
& E & 77 & 6 & 12 & 15 \\
& S & 37 & 15 & 22 & 36 \\
& G & 53 & 23 & 19 & 15 \\
\bottomrule
\end{tabular}
\end{table}

\section{Methodology for ESG modelling}
Our methods used to classify ESG-related sentiment on the proposed dataset focus on two different perspectives: adapting pre-trained machine learning models (such as BERT and TabPFN) and zero-shot querying of LLMs, ranging from the monolingual Slovene model GaMS to the multilingual reasoning model GPT-OSS, as well as building a hirearchically stacked ESG model.

\begin{figure}
    \centering
    \includegraphics[width=0.5\textwidth]{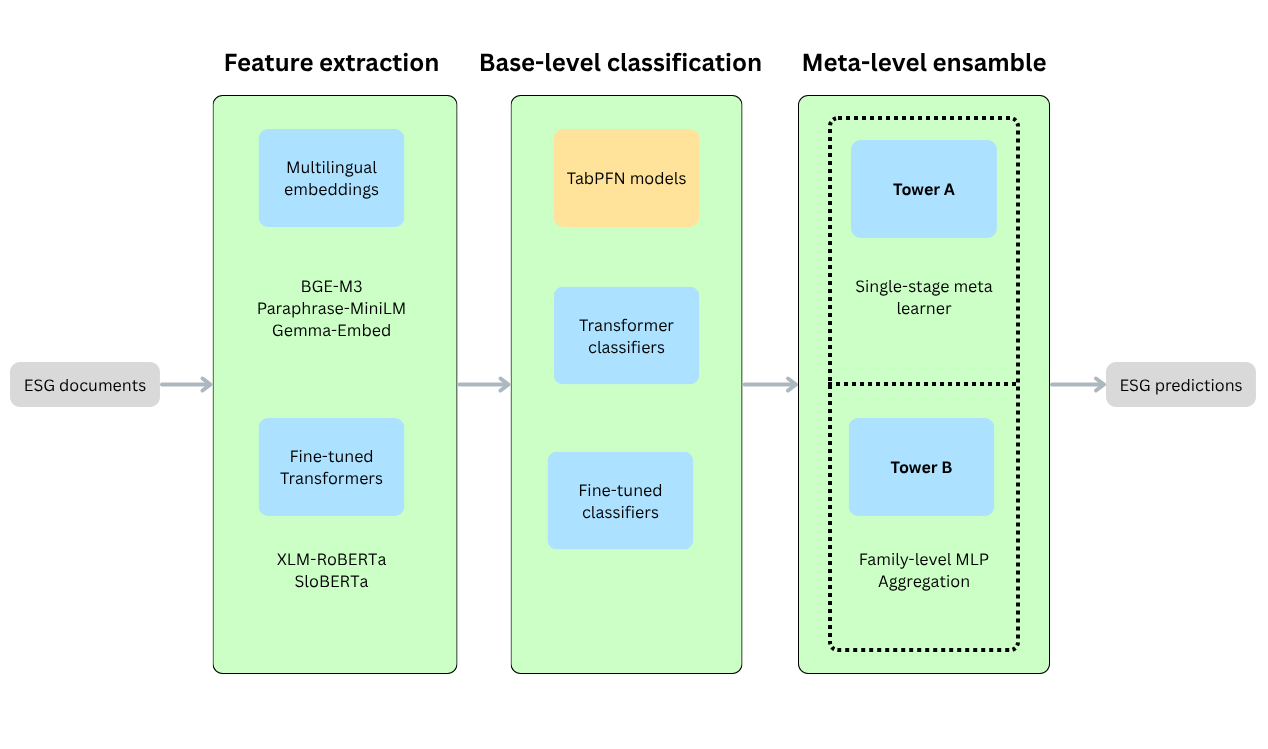}
    \caption{Methodology pipeline}
    \label{fig:architecture}
\end{figure}

Our approach follows a multi-level stacking paradigm consisting of three principal stages: a) feature extraction through multiple text representation methods, b) base-level classification using diverse model families, and c) meta-level prediction through hierarchical neural ensembles.
  The complete pipeline is illustrated in Figure~\ref{fig:architecture}.

\subsection{Text Representation Models}

We employ five distinct text encoding strategies.

\subsubsection{Multilingual Sentence Embeddings}

\textbf{BGE-M3}\footnote{\url{https://huggingface.co/BAAI/bge-m3}} (BAAI/bge-m3): A state-of-the-art multilingual embedding model supporting over 100 languages. The model produces dense 1024-dimensional vectors optimized for semantic similarity tasks through contrastive learning on large-scale multilingual corpora. 

\textbf{Paraphrase-Multilingual-MiniLM-L12-v2}\footnote{\url{https://huggingface.co/sentence-transformers/paraphrase-multilingual-MiniLM-L12-v2}}: A distilled sentence transformer architecture based on the MiniLM framework \cite{wang2020minilm}, providing computationally efficient 384-dimensional embeddings. 

\textbf{Gemma-Embed}\footnote{\url{https://huggingface.co/google/embeddinggemma-300m}} (google/embeddinggemma-300m): A task-agnostic embedding model built on Google's Gemma architecture \citep{team2024gemma}. 

\subsubsection{Fine-tuned Transformer Classifiers}

\textbf{XLM-RoBERTa-base}~\cite{conneau2020unsupervised}: A cross-lingual pre-trained transformer model trained on 2.5TB of CommonCrawl data covering 100 languages. Unlike multilingual BERT, XLM-RoBERTa employs no language-specific embeddings, instead learning cross-lingual representations through language-agnostic pretraining. We fine-tune all layers on the ESG classification task with a linear classification head (768 $\rightarrow$ 4 classes per aspect).

\textbf{SloBERTa}~\citep{ulcar2021sloberta}: A RoBERTa variant specifically pre-trained on Slovenian texts. This model provides specialized morphological and syntactic knowledge for Slovenian, a highly inflected South Slavic language. The architecture mirrors RoBERTa-base with language-specific tokenization and vocabulary.

\subsubsection{Dimensionality Reduction}

To address TabPFN's computational constraints on high-dimensional inputs, we optionally apply Truncated Singular Value Decomposition (SVD) to the embedding matrices. We evaluate candidate dimensions $\mathcal{D} = \{$32, 64, 128, 256$\}$ through nested validation, selecting the dimensionality $d^* \in \mathcal{D}$ that maximizes macro-averaged F1 score on a held-out 20\% internal validation split from the training partition.

\subsection{Base-Level Classification}
The assign positive, negative, neutral or irrelevant label for each ESG category.
\subsubsection{TabPFN-based Models}

TabPFN (Prior-Fitted Networks) \citep{hollmann2022tabpfn} is a meta-learned classifier that performs approximate Bayesian inference through in-context learning without gradient-based training. Given embedding matrix $\mathbf{X} \in \mathbb{R}^{n \times d}$ and labels $\mathbf{y}$, TabPFN produces probabilistic predictions $p(\mathbf{y}^* | \mathbf{X}, \mathbf{y}, \mathbf{x}^*)$ for test instances $\mathbf{x}^*$ through a single forward pass, leveraging patterns learned from synthetic tabular datasets during meta-training.

We construct six TabPFN-based classifiers by pairing three embedding models (BGE-M3, Paraphrase-MiniLM, Gemma-Embed) with two preprocessing variants (with/without SVD). Each classifier operates independently on the E, S, and G aspects, producing 4-class probability distributions. Model identifiers follow the convention \texttt{tabpfn\_\{embedding\}} and \texttt{tabpfn\_\{embedding\}\_svd}.

\subsubsection{Fine-tuned Transformer Models}

Transformer models are trained as sequence classification systems using the Hugging Face Trainer API \cite{wolf2020transformers} with the hyper-parameters specified in Table~\ref{tab:hf_hyperparams}.  Each aspect ($a \in \{\text{E}, \text{S}, \text{G}\}$) is trained independently as a 4-class classification task (pos, neg, neut, irr), resulting in aspect-specific fine-tuned models. Model identifiers follow the convention \texttt{hf\_(SloBERTa/XLMR)}.

\begin{table}[t]
\centering
\small
\begin{tabular}{ll}
\toprule
\textbf{Hyperparameter} & \textbf{Value} \\
\midrule
Optimizer & AdamW \\
Learning rate & $3 \times 10^{-5}$ \\
Weight decay & 0.01 \\
Batch size & 128 \\
Max sequence length & 192 tokens \\
Warmup steps & 100 \\
Learning rate schedule & Linear \\
Max epochs & 100 \\
Early stopping patience & 15 epochs \\
Metric for model selection & Macro-F1 \\
\bottomrule
\end{tabular}
\caption{Hyperparameters for transformer fine-tuning.}
\label{tab:hf_hyperparams}
\end{table}

\subsection{Large Language Models}

To assess the performance of instruction-following LLMs in zero-shot and few-shot ESG sentiment classification, we evaluate five models from the Gemma and GPT families. Unlike fine-tuned transformers, these models are prompted to classify ESG sentiment without gradient-based adaptation. We employ a structured prompt template that presents the classification task with explicit ESG definitions and class descriptions. We evaluate models of varying parameter counts to assess the impact of scale on ESG classification:

\begin{itemize}
    \item \textbf{GaMS-9B / GaMS-27B}~\cite{gams}: Gemma-based models fine-tuned for Slovenian language understanding
    \item \textbf{Gemma3-12B / Gemma3-27B}~\citep{team2024gemma}: Instruction-tuned variants from the Gemma 3 family
    \item \textbf{gpt-oss 20B}~\cite{Agarwal2025gptoss120bG}: An open-source GPT-architecture reasoning model with 20B parameters
\end{itemize}

\textbf{Inference Configuration}: For certain models, we explore few-shot prompting by including $k \in \{10, 20\}$ labeled examples in the prompt context (denoted by model suffix, e.g., Gemma3-12B). Temperature is set to 0.0 for deterministic outputs, and responses are parsed to extract class predictions for each ESG aspect.

LLMs are evaluated directly on the test set $\mathcal{D}_{\text{test}}$ without additional training, providing a comparison baseline for zero-shot transfer performance against fine-tuned and ensemble approaches.

\subsection{Meta-Feature Construction}

Base model predictions are transformed into meta-features through the following pipeline:

\begin{enumerate}
    \item \textbf{Probability Extraction}: Each base model produces probability distributions $\mathbf{P}_a \in \mathbb{R}^{n \times 4}$ for aspect $a \in \{\text{E}, \text{S}, \text{G}\}$
    
    \item \textbf{Logit Transformation}: Convert probabilities to logits to handle extreme values and provide unbounded feature space:
    \begin{equation}
    \mathbf{L}_a = \log(\text{clip}(\mathbf{P}_a, \epsilon, 1.0))
    \label{eq:logit_transform}
    \end{equation}
    where $\epsilon = 10^{-6}$ prevents numerical instability.
    
    \item \textbf{Concatenation}: For each base model, concatenate aspect logits:
    \begin{equation}
    \mathbf{X}_{\text{base}} = [\mathbf{L}_{\text{E}} \,||\, \mathbf{L}_{\text{S}} \,||\, \mathbf{L}_{\text{G}}] \in \mathbb{R}^{n \times 12}
    \label{eq:meta_concat}
    \end{equation}
\end{enumerate}

This transformation preserves relative probability magnitudes while providing a more stable feature space for meta-learning, avoiding the compression of probabilities near 0 or 1 that can occur in linear scaling.

\subsection{Meta-Level Ensemble Architecture}

We propose two hierarchical ensemble strategies that differ in their aggregation topology.


\textbf{Tower A} employs a single-stage meta-learner that processes concatenated meta-features from all selected base families:
\begin{equation}
\mathbf{X}_{\text{meta}} = [\mathbf{X}_{\text{fam}_1} \,||\, \mathbf{X}_{\text{fam}_2} \,||\, \cdots \,||\, \mathbf{X}_{\text{fam}_k}] \in \mathbb{R}^{n \times 12k}
\label{eq:tower_a}
\end{equation}

\noindent where $k$ is the number of base model families. This architecture allows the meta-learner to discover arbitrary cross-family interaction patterns.


\textbf{Tower B} implements a two-level hierarchy to exploit family-specific characteristics:
\begin{enumerate}
    \item \textbf{Level 1 (Family-Specific Meta-Models)}: Each base family $i$ trains an independent meta-MLP:
    \begin{equation}
        \mathbf{Z}_i = \text{MLP}_{\text{fam}_i}(\mathbf{X}_{\text{fam}_i}) \in \mathbb{R}^{n \times 12}
        \label{eq:tower_b_level1}
    \end{equation}

    \item \textbf{Level 2 (Cross-Family Aggregation)}: A second meta-MLP combines family-level outputs:
    \begin{equation}
        \hat{\mathbf{Y}} = \text{MLP}_{\text{final}}([\mathbf{Z}_1 \,||\, \mathbf{Z}_2 \,||\, \cdots \,||\, \mathbf{Z}_k])
        \label{eq:tower_b_level2}
    \end{equation}
\end{enumerate}

This architecture allows each family to learn specialized combination strategies (e.g., TabPFN families may benefit from uncertainty calibration while transformer families may require confidence rescaling) before global aggregation.

\textbf{Meta-MLP Architecture} All meta-models share a unified neural architecture with multi-task learning formulation:
\begin{align}
\mathbf{h}^{(1)} &= \text{ReLU}(\text{BN}(\mathbf{W}^{(1)}\mathbf{x} + \mathbf{b}^{(1)})) \label{eq:mlp_layer1} \\
\mathbf{h}^{(1)}_{\text{drop}} &= \text{Dropout}(\mathbf{h}^{(1)}, p=0.4) \label{eq:mlp_dropout} \\
\mathbf{z} &= \text{ReLU}(\mathbf{W}^{(2)}\mathbf{h}^{(1)}_{\text{drop}} + \mathbf{b}^{(2)}) \label{eq:mlp_shared}
\end{align}

\noindent where $\mathbf{W}^{(1)} \in \mathbb{R}^{64 \times d_{\text{in}}}$, $\mathbf{W}^{(2)} \in \mathbb{R}^{64 \times 64}$, and BN denotes batch normalization. The shared trunk $\mathbf{z}$ feeds into three aspect-specific prediction heads:
\begin{equation}
\mathbf{o}_a = \mathbf{W}^{(3)}_a \text{ReLU}(\mathbf{W}^{(2)}_a \mathbf{z} + \mathbf{b}^{(2)}_a) + \mathbf{b}^{(3)}_a
\label{eq:mlp_heads}
\end{equation}

\noindent for $\quad a \in \{\text{E}, \text{S}, \text{G}\}$, where $\mathbf{W}^{(2)}_a \in \mathbb{R}^{32 \times 64}$ and $\mathbf{W}^{(3)}_a \in \mathbb{R}^{4 \times 32}$.

The multi-task formulation with shared representations encourages learning of correlations between ESG aspects (e.g., environmental practices often correlate with governance structures) while maintaining aspect-specific prediction capacity.

\textbf{Loss Function}: Joint cross-entropy across all aspects:
\begin{equation}
\mathcal{L} = \text{CE}(\mathbf{o}_{\text{E}}, \mathbf{y}_{\text{E}}) + \text{CE}(\mathbf{o}_{\text{S}}, \mathbf{y}_{\text{S}}) + \text{CE}(\mathbf{o}_{\text{G}}, \mathbf{y}_{\text{G}})
\label{eq:meta_loss}
\end{equation}

\textbf{Optimization}: AdamW with learning rate $10^{-3}$, weight decay 0.01, batch size 64.

\subsubsection{Training Protocol: Stratified 80/20 Split}

We employ a stratified holdout protocol to balance computational efficiency with robust evaluation. The training procedure consists of three stages:

\textbf{Stage 1: Data Partitioning} The training corpus $\mathcal{D}_{\text{train}}$ is partitioned into training ($\mathcal{D}_{80}$) and validation ($\mathcal{D}_{20}$) subsets using stratified sampling. To preserve the joint distribution of ESG labels, we implement multilabel stratification on the (E, S, G) triplets using iterative stratification \cite{sechidis2011stratification}. This ensures that the validation set maintains representative samples from all 64 possible ESG label combinations (4 $\times$ 4 $\times$ 4), preventing evaluation bias from rare triplet configurations.

\textbf{Stage 2: Base Model Training}

Each base model family is trained exclusively on $\mathcal{D}_{80}$ and generates predictions on both $\mathcal{D}_{20}$ (validation) and $\mathcal{D}_{\text{test}}$ (held-out test set):

\begin{enumerate}
    \item \textbf{Embedding Models + TabPFN}: Extract embeddings from $\mathcal{D}_{80}$, optionally apply SVD dimensionality reduction, fit TabPFN classifier, predict on $\mathcal{D}_{20}$ and $\mathcal{D}_{\text{test}}$
    
    \item \textbf{Transformer Models}: Fine-tune on $\mathcal{D}_{80}$ with early stopping based on $\mathcal{D}_{20}$ performance, generate final predictions on $\mathcal{D}_{20}$ and $\mathcal{D}_{\text{test}}$ using best checkpoint
\end{enumerate}

This produces two sets of meta-features per base model:
\begin{itemize}
    \item $\mathbf{X}_{\text{meta}}^{(20)} \in \mathbb{R}^{|\mathcal{D}_{20}| \times 12}$: Meta-features for validation samples
    \item $\mathbf{X}_{\text{meta}}^{(\text{test})} \in \mathbb{R}^{|\mathcal{D}_{\text{test}}| \times 12}$: Meta-features for test samples
\end{itemize}

\textbf{Stage 3: Meta-Model Training.}
Meta-models are trained on $\mathbf{X}{\text{meta}}^{(20)}$ with early stopping: split $\mathcal{D}{20}$ into $80\%/20\%$ (stratified) meta-train/validation, train up to $200$ epochs while tracking validation loss, select $t^*$ with minimum validation loss (patience=15), retrain on all of $\mathcal{D}_{20}$ for $t^*$ epochs, then generate final predictions on $\mathcal{D}_{\text{test}}$. This nested validation promotes generalization and reduces overfitting to base-model biases.


To ensure robustness against random initialization effects, we repeat the entire pipeline across three independent random seeds $\mathcal{S} = \{0, 100, 200\}$.

\subsection{Evaluation Metrics}

Model performance is assessed using four complementary metrics, computed independently for each ESG aspect:

\begin{itemize}
    \item \textbf{Accuracy}: $\text{Acc} = \frac{1}{n}\sum_{i=1}^{n} \mathbb{1}[\hat{y}_i = y_i]$
    
    \item \textbf{Macro-averaged F1}: Harmonic mean of precision and recall across classes without class-weighting:
    \begin{equation}
    \text{F1}_{\text{macro}} = \frac{1}{C} \sum_{c=1}^{C} \frac{2 \cdot \text{Prec}_c \cdot \text{Rec}_c}{\text{Prec}_c + \text{Rec}_c}
    \label{eq:macro_f1}
    \end{equation}
    
    \item \textbf{Balanced Accuracy}: Arithmetic mean of per-class recall, accounting for class imbalance:
    \begin{equation}
    \text{BAcc} = \frac{1}{C} \sum_{c=1}^{C} \frac{\text{TP}_c}{\text{TP}_c + \text{FN}_c}
    \label{eq:balanced_acc}
    \end{equation}
    
    \item \textbf{Area Under Precision-Recall Curve (AUPRC)}: Average precision across one-vs-rest binary decompositions, providing a single-number summary of the precision-recall trade-off
\end{itemize}

System-level performance is reported as the mean across aspects and seeds, with standard deviation indicating inter-seed variability.



\section{Results and Discussion}

The results presented in Tables \ref{tab:test_results_E}–\ref{tab:test_results_G} provide clear evidence that transformer-based architectures, supported by ensemble and multi-task learning strategies, are well suited for ESG sentiment classification in Slovene news. The consistent performance gains achieved by the multi-task fusion models across all three ESG dimensions indicate that the aspects of Environmental, Social, and Governance sentiment share underlying linguistic cues that can be effectively captured through shared representations. This interdependence highlights that public discourse around ESG topics is often contextually entangled—positive environmental narratives tend to correlate with favorable governance and social framing, and vice versa.

The Environmental aspect (Table \ref{tab:test_results_E}) shows the strongest overall results, with macro-F1 values surpassing 0.6 for the top-performing models. The superior accuracy of the ensemble “Final Tower” architectures suggests that aggregating diverse feature spaces—sentence embeddings, fine-tuned transformer outputs, and meta-learned representations—yields a more comprehensive understanding of ESG-related sentiment. Notably, the SloBERTa model achieves robust scores comparable to or exceeding multilingual alternatives, confirming that monolingual pretraining remains advantageous for highly inflected languages such as Slovene. By contrast, multilingual models like XLM-RoBERTa exhibit more stable but less specialized behavior, implying that language-agnostic pretraining can miss subtle morphological or idiomatic sentiment signals present in the Slovene media corpus.
Performance for the Social aspect (Table \ref{tab:test_results_S}) is comparatively lower, with macro-F1 values clustering between 0.30 and 0.45 across models. This can be attributed to the abstract and context-sensitive nature of social issues—topics like labor relations or equality often rely on nuanced framing rather than explicit sentiment markers. Interestingly, the multilingual models performed competitively in this category, suggesting that cross-lingual exposure may help recognize generalized social discourse patterns. The relative underperformance of ensemble systems in this dimension further supports the idea that social sentiment requires more contextual or pragmatic interpretation than currently encoded by the models. 
The Governance aspect (Table \ref{tab:test_results_G}) remains the most challenging dimension, reflected in lower average macro-F1 values and wider variance between seeds. This weakness likely stems from ambiguity in annotator interpretations and the abstract, institutional tone typical of governance reporting. Governance language often lacks clear evaluative expressions, making sentiment polarity difficult to infer even for human annotators. The observed correspondence between lower inter-annotator agreement and reduced model performance supports this interpretation and underlines the difficulty of operationalizing governance sentiment in textual data.

Overall, the results validate the study’s design choices while also exposing limitations inherent to ESG text analysis in news. The small dataset size (550 annotated articles) constrains generalization, particularly for multi-class classification across three interrelated sentiment axes. Moreover, the reliance on LLM-assisted filtering introduces potential sampling bias, as model-based preselection may favor easily classifiable or lexically explicit texts. The ESG sentiment categories themselves may overlap semantically, challenging both human and machine annotation consistency. Future studies could mitigate these issues through larger, more balanced datasets and clearer annotation guidelines emphasizing cross-aspect distinctions.




\begin{table}[t]
\centering
\caption{Test-set results for Aspect E (mean over seeds). Primary metric is F1-macro; we also report AUPRC, Balanced Accuracy (BAcc), and Accuracy. Best results per column are in bold.}
\label{tab:test_results_E}
\resizebox{\linewidth}{!}{\begin{tabular}{lcccc}
\toprule
\textbf{Model} & \textbf{Accuracy} & \textbf{F1-macro} & \textbf{BAcc} & \textbf{AUPRC} \\
\midrule
\multicolumn{5}{l}{\emph{Baseline}} \\
Majority & 0.7000 & 0.2059 & 0.2500 & 0.2500 \\
\midrule
\multicolumn{5}{l}{\emph{Ensemble}} \\
FinalTowerA & 0.7394 & 0.4469 & 0.5010 & 0.5638 \\
FinalTowerB & 0.7182 & 0.4291 & 0.4476 & 0.4815 \\
\midrule
\multicolumn{5}{l}{\emph{Fine-tuned Transformers}} \\
hf\_sloberta & 0.7242 & 0.4284 & 0.4648 & 0.5166 \\
hf\_xlm-roberta & 0.7212 & 0.4251 & 0.4572 & 0.5260 \\
\midrule
\multicolumn{5}{l}{\emph{Sentence-Transformer}} \\
tabpfn\_bge-m3 & 0.7455 & 0.3841 & 0.3986 & 0.5198 \\
tabpfn\_gemma-embed & 0.7182 & 0.3402 & 0.3412 & 0.4404 \\
tabpfn\_paraphrase & 0.7091 & 0.3691 & 0.3972 & 0.4548 \\
\midrule
\multicolumn{5}{l}{SVD} \\
tabpfn\_bge-m3\_svd & 0.7727 & 0.4717 & 0.4648 & 0.5847 \\
tabpfn\_gemma-embed\_svd & 0.7212 & 0.3972 & 0.3995 & 0.4894 \\
tabpfn\_paraphrase\_svd & 0.7424 & 0.3970 & 0.4103 & 0.5107 \\
\midrule
\multicolumn{5}{l}{\emph{LLMs}} \\
GaMS-27B 10 & 0.8000 & 0.5108 & 0.5102 & 0.4441 \\
GaMS-9B & 0.7545 & 0.4255 & 0.3926 & 0.3484 \\
Gemma3-12B & 0.7818 & 0.5375 & 0.5449 & 0.4546 \\
Gemma3-27B & 0.7818 & \textbf{0.6106} & \textbf{0.6777} & 0.4895 \\
gpt-oss 20B& \textbf{0.8182} & 0.5907 & 0.5828 & \textbf{0.5847} \\
\bottomrule
\end{tabular}}
\end{table}

\begin{table}[t]
\centering
\caption{Test-set results for Aspect S (mean over seeds). Primary metric is F1-macro; we also report AUPRC, Balanced Accuracy (BAcc), and Accuracy. Best results per column are in bold.}
\label{tab:test_results_S}
\resizebox{\linewidth}{!}{\begin{tabular}{lcccc}
\toprule
\textbf{Model} & \textbf{Accuracy} & \textbf{F1-macro} & \textbf{BAcc} & \textbf{AUPRC} \\
\midrule
\multicolumn{5}{l}{\emph{Baseline}} \\
Majority & 0.3364 & 0.1259 & 0.2500 & 0.2500 \\
\midrule
\multicolumn{5}{l}{\emph{Ensemble}} \\
FinalTowerA & 0.4606 & 0.3033 & 0.3589 & 0.3515 \\
FinalTowerB & 0.4364 & 0.3252 & 0.3602 & 0.3557 \\
\midrule
\multicolumn{5}{l}{\emph{Fine-tuned Transformers}} \\
hf\_sloberta & 0.4909 & 0.4238 & 0.4265 & 0.4475 \\
hf\_xlm-roberta & 0.4939 & 0.4404 & 0.4423 & \textbf{0.4533} \\
\midrule
\multicolumn{5}{l}{\emph{Sentence-Transformer}} \\
tabpfn\_bge-m3 & 0.4455 & 0.2726 & 0.3360 & 0.4051 \\
tabpfn\_gemma-embed & 0.4515 & 0.3177 & 0.3529 & 0.4083 \\
tabpfn\_paraphrase & 0.4424 & 0.2674 & 0.3336 & 0.4072 \\
\midrule
\multicolumn{5}{l}{SVD} \\
tabpfn\_bge-m3\_svd & 0.4515 & 0.2804 & 0.3418 & 0.3821 \\
tabpfn\_gemma-embed\_svd & 0.4727 & 0.2940 & 0.3600 & 0.3962 \\
tabpfn\_paraphrase\_svd & 0.4545 & 0.2804 & 0.3443 & 0.3987 \\
\midrule
\multicolumn{5}{l}{\emph{LLMs}} \\
GaMS-27B & 0.5000 & 0.4140 & 0.4358 & 0.3395 \\
GaMS-9B & 0.5182 & 0.4193 & 0.4432 & 0.3529 \\
Gemma3-12B & 0.4182 & 0.3717 & 0.4179 & 0.3369 \\
Gemma3-27B & 0.4455 & 0.4022 & \textbf{0.4622} & 0.3430 \\
gpt-oss 20B& \textbf{0.5273} & \textbf{0.4512} & 0.4547 & 0.3603 \\
\bottomrule
\end{tabular}}
\end{table}

\begin{table}[t]
\centering
\small
\caption{Test-set results for Aspect G (mean over seeds). Primary metric is F1-macro; we also report AUPRC, Balanced Accuracy (BAcc), and Accuracy. Best results per column are in bold.}
\label{tab:test_results_G}
\resizebox{\linewidth}{!}{\begin{tabular}{lcccc}
\toprule
\textbf{Model} & \textbf{Accuracy} & \textbf{F1-macro} & \textbf{BAcc} & \textbf{AUPRC} \\
\midrule
\multicolumn{5}{l}{\emph{Baseline}} \\
Majority & 0.4818 & 0.1626 & 0.2500 & 0.2500 \\
\midrule
\multicolumn{5}{l}{\emph{Ensemble}} \\
FinalTowerA & 0.5152 & 0.3742 & 0.3997 & 0.4351 \\
FinalTowerB & 0.4939 & 0.3935 & 0.4158 & 0.4342 \\
\midrule
\multicolumn{5}{l}{\emph{Fine-tuned Transformers}} \\
hf\_sloberta & 0.6091 & \textbf{0.5420} & \textbf{0.5486} & \textbf{0.5528} \\
hf\_xlm-roberta & 0.5333 & 0.3590 & 0.3843 & 0.4849 \\
\midrule
\multicolumn{5}{l}{\emph{Sentence-Transformer}} \\
tabpfn\_bge-m3 & 0.5879 & 0.4457 & 0.4571 & 0.5164 \\
tabpfn\_gemma-embed & 0.5636 & 0.3867 & 0.3974 & 0.4751 \\
tabpfn\_paraphrase & 0.5818 & 0.4239 & 0.4337 & 0.4739 \\
\midrule
\multicolumn{5}{l}{SVD} \\
tabpfn\_bge-m3\_svd & \textbf{0.6879} & 0.5210 & 0.5441 & 0.5386 \\
tabpfn\_gemma-embed\_svd & 0.6030 & 0.3909 & 0.4248 & 0.4867 \\
tabpfn\_paraphrase\_svd & 0.5788 & 0.4119 & 0.4349 & 0.4560 \\
\midrule
\multicolumn{5}{l}{\emph{LLMs}} \\
GaMS-27B & 0.5000 & 0.3899 & 0.4405 & 0.3620 \\
GaMS-9B & 0.6000 & 0.4452 & 0.4591 & 0.3730 \\
Gemma3-12B & 0.4727 & 0.4870 & 0.5120 & 0.4294 \\
Gemma3-27B & 0.4091 & 0.4146 & 0.4529 & 0.3816 \\
gpt-oss 20B& 0.5364 & 0.4792 & 0.4821 & 0.3839 \\
\bottomrule
\end{tabular}}
\end{table}

\section{Case Study: Qualitative Temporal ESG Evaluation}

After evaluating the proposed models, we select the gpt-oss-20b model to analyze the sentiment distribution over time for four companies of interest by analysing a large news media monitoring dataset for the period 2010-2025. The annual average sentiment score is computed by subtracting the count of negative sentiment articles from the count of positive sentiment articles. 

\begin{table}[!h]
\centering
\caption{ESG Sentiment Analysis Summary by Company and Category}
\label{tab:esg_summary}
\resizebox{\linewidth}{!}{\begin{tabular}{llcccccc}
\toprule
\textbf{Company} & \textbf{Category} & \textbf{Total} & \textbf{Relevant} & \textbf{Positive} & \textbf{Negative} & \textbf{Neutral} & \textbf{Irrelevant} \\
\midrule
Talum & E & 4072 & 1234 & 460 & 448 & 326 & 2838 \\
 & S &  & 2446 & 952 & 900 & 594 & 1626 \\
 & G &  & 2404 & 512 & 1080 & 812 & 1668 \\
\midrule
Sdh & E & 30338 & 2504 & 668 & 838 & 998 & 27834 \\
 & S &  & 15758 & 1892 & 7908 & 5958 & 14580 \\
 & G &  & 27640 & 2082 & 14610 & 10948 & 2698 \\
\midrule
Cinkarna & E & 11062 & 1516 & 364 & 794 & 358 & 9546 \\
 & S &  & 2754 & 666 & 1178 & 910 & 8308 \\
 & G &  & 3502 & 418 & 1696 & 1388 & 7560 \\
\midrule
Salonit & E & 8026 & 1408 & 356 & 862 & 190 & 6618 \\
 & S &  & 1810 & 464 & 970 & 376 & 6216 \\
 & G &  & 1970 & 234 & 1182 & 554 & 6056 \\
\bottomrule
\end{tabular}}
\end{table}

\begin{figure}[h!]
    \centering
    \resizebox{\linewidth}{!}{\includegraphics{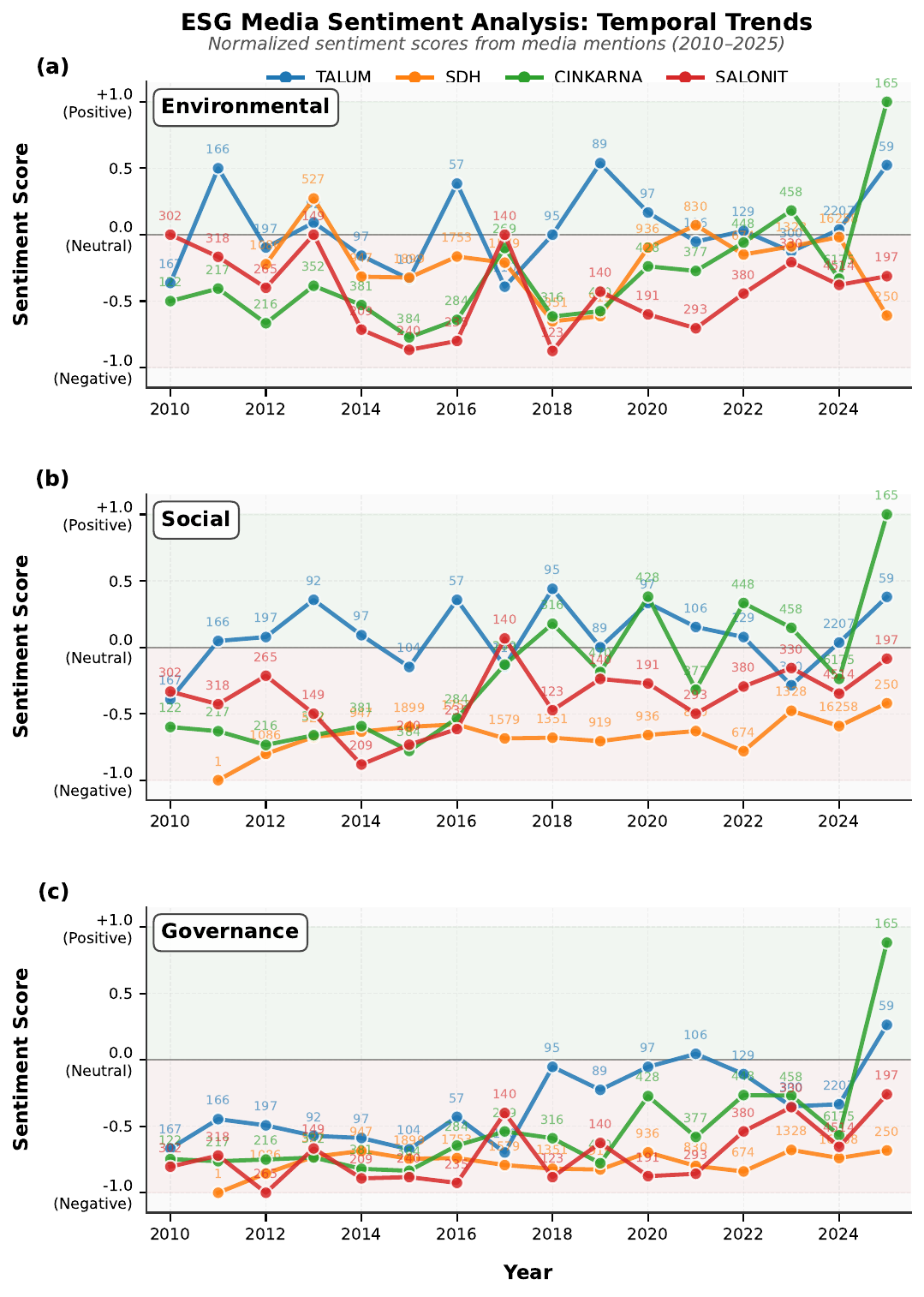}}
    \caption{Normalized sentiment scores from media mentions (2010 2025)}
    \label{fig:placeholder}
\end{figure}

These companies were selected as representative cases of different approaches to sustainability, governance, and community engagement within Slovenian industry. \textit{Talum }exemplifies a successful transition from a high–environmental-impact aluminum producer to a recycling-based model, effectively balancing environmental responsibility with its role as a major regional employer. Similarly, \textit{SDH}, as a state holding company managing publicly owned enterprises, has introduced high governance standards and driven the adoption of ESG reporting across state-managed firms, consistent with research suggesting that government ownership often fosters sustainability commitments \citep{qian-yang}.

\textit{Cinkarna Celje }reflects a long-term transformation from a historically polluting zinc producer to a company committed to environmental remediation and local well-being, actively monitoring soil conditions and the health of nearby residents. In contrast, \textit{Anhovo / Alpacem} illustrates the social tensions that can arise when industrial development and community interests diverge. Once associated with asbestos production and its severe societal impacts, the company’s more recent plans to expand into waste incineration have sparked public resistance due to uncertainties about environmental and health consequences. Together, the cases capture a spectrum of corporate responses to sustainability pressures—from proactive adaptation and transparency to ongoing conflict and mistrust.

In this case study, an economic expert compared temporal ESG-sentiment analysis with key business events as described in relevant CEO letters for Cinkarna Celje and Talum.

At Cinkarna Celje, ESG sentiment patterns indicate strong sensitivity to regulatory and governance-related events. The announcement of environmental remediation in 2017 led to a significant improvement in E and S sentiment, while the subsequent slow remediation process, coupled with a lawsuit by the European Commission over delays in closing the landfills and the question at the EU level regarding whether the raw material for core product, titanium dioxide, is carcinogenic, affected sentiment. G sentiment increased during board and management changes in 2020 and again in 2025 with the reappointment of the same CEO, but declined in 2024 when a board member resigned, suggesting that leadership stability and continuity are positively valued. Governance sentiment also correlates positively with dividend payments, reflecting an association between shareholder returns and perceptions of governance quality. The exceptionally high sentiment in 2025 coincides with the European Court of Justice ruling that titanium dioxide is not carcinogenic – a decision with limited impact on actual environmental and health outcomes, but significant reputational impact, illustrating how institutional signals can reshape ESG perception independently of environmental performance. At Talum in 2015, financial results turned positive after a prolonged period of losses: strategic goals were achieved, the main shareholder increased its equity investment, and employees were highly engaged in innovative processes. However, sentiment in all three pillars (E, S, and G) declined, suggesting persistent scepticism despite improved financial performance. Sentiment rebounded in 2016, supported by strategic restructuring, innovation, and workforce expansion, but fell again in 2017 as environmental sentiment weakened despite the company’s continued commitment to efficient and sustainable production, an increased workforce, and doubled profit. Between 2018 and 2019, E sentiment strengthened as Talum invested in restructuring its production towards carbon-neutral products with high added value, while S sentiment declined due to perceived risks to employment associated with the reduction of primary aluminium production. In 2025, all three sentiment dimensions were strongly positive, reflecting the completion of Talum’s green transformation, technological modernisation, and diversification into new industries such as commerce, pharmaceuticals, and defence. It is notable that investments in defence seem no longer to be considered ESG “problematic” in 2025, probably due to the geopolitical situation. Both firms exhibited markedly positive S sentiment in 2020, coinciding with the COVID-19 pandemic. This increase is likely related to the companies’ ability to maintain stable operations and retain employees despite disrupted market conditions, reinforcing employment security as a key driver of social sentiment. Across both companies, announcements and disbursements of employee bonuses consistently coincided with positive shifts in S sentiment, suggesting that distributive and welfare-related actions have a measurable influence on social evaluations. Workforce contraction had limited effect on S sentiment in Cinkarna, whereas in Talum, S sentiment is highly sensitive to any possible impact of any of the conditions on employment. Across the two companies, ESG sentiment is only weakly related to financial indicators (profit, liquidity, efficiency). Instead, communicative and institutional factors - regulatory decisions, board changes, and employee-related gestures - exert a stronger and more immediate effect. These findings indicate that text-based ESG sentiment primarily reflects the social construction of corporate responsibility rather than direct economic or environmental outcomes.
\vspace{-1em}

\section{Conclusions and Further Work}
This work presents the first publicly available Slovene ESG dataset and uses it as a resource for training  LLM-based, transformer-based classification models and hierarchical ensembling. Beyond technical performance, these findings have broader implications for sustainability analytics: automated monitoring of ESG sentiment could provide dynamic, fine-grained insights into corporate reputation shifts across time and media outlets. Our results show that LLMs lead Environmental (Gemma3-27B, F1-macro 0.61) and Social (gpt-oss 20B, 0.45) tasks, while fine-tuned SloBERTa tops Governance (0.54).
Future research should pursue several directions. Expanding the dataset temporally and thematically would enhance robustness. Incorporating temporal and causal modeling could capture how specific events—policy changes, environmental incidents, or governance scandals—affect sentiment trajectories. The most interesting line of research is to observe the ESG assigned sentiment in relation to ESG financial information.

\section{Code Availability}
The source code is publicly available at
\url{https://github.com/bkolosk1/slo-news-esg}.

\section{Data Availability}
The dataset is publicly available at
\url{http://hdl.handle.net/11356/2102}.

\section*{Acknowledgments}

This work was supported by the Slovenian Research and Innovation Agency (ARIS) through the projects EMMA (Embeddings-based Techniques for Media Monitoring Applications; L2-50070), Large Language Models for Digital Humanities (LLM4DH; GC-0002), and the research core funding programme Knowledge Technologies (P2-0103). BK is supported by the Young Researcher Grant PR-12394.

\section{Bibliographical References}\label{sec:reference}

\bibliographystyle{lrec2026-natbib}
\bibliography{lrec2026-example}

@article{chen2023environmental,
  title={Environmental, social, and governance (ESG) performance and financial outcomes: Analyzing the impact of ESG on financial performance},
  author={Chen, Simin and Song, Yu and Gao, Peng},
  journal={Journal of environmental management},
  volume={345},
  pages={118829},
  year={2023},
  publisher={Elsevier}
}

@inproceedings{ivacic-etal-2023-analysis,
    title = "Analysis of Transfer Learning for Named Entity Recognition in {S}outh-{S}lavic Languages",
    author = "Iva{\v{c}}i{\v{c}}, Nikola  and
      Tran, Thi Hong Hanh  and
      Koloski, Boshko  and
      Pollak, Senja  and
      Purver, Matthew",
    editor = "Piskorski, Jakub  and
      Marci{\'n}czuk, Micha{\l}  and
      Nakov, Preslav  and
      Ogrodniczuk, Maciej  and
      Pollak, Senja  and
      P{\v{r}}ib{\'a}{\v{n}}, Pavel  and
      Rybak, Piotr  and
      Steinberger, Josef  and
      Yangarber, Roman",
    booktitle = "Proceedings of the 9th Workshop on Slavic Natural Language Processing 2023 (SlavicNLP 2023)",
    month = may,
    year = "2023",
    address = "Dubrovnik, Croatia",
    publisher = "Association for Computational Linguistics",
    url = "https://aclanthology.org/2023.bsnlp-1.13/",
    doi = "10.18653/v1/2023.bsnlp-1.13",
    pages = "106--112"
}

@inproceedings{non-etal-2022-macocu,
    title = "{M}a{C}o{C}u: Massive collection and curation of monolingual and bilingual data: focus on under-resourced languages",
    author = "Ba{\~n}{\'o}n, Marta  and
      Espl{\`a}-Gomis, Miquel  and
      Forcada, Mikel L.  and
      Garc{\'i}a-Romero, Cristian  and
      Kuzman, Taja  and
      Ljube{\v{s}}i{\'c}, Nikola  and
      van Noord, Rik  and
      Sempere, Leopoldo Pla  and
      Ram{\'i}rez-S{\'a}nchez, Gema  and
      Rupnik, Peter  and
      Suchomel, V{\'i}t  and
      Toral, Antonio  and
      van der Werff, Tobias  and
      Zaragoza, Jaume",
    editor = {Moniz, Helena  and
      Macken, Lieve  and
      Rufener, Andrew  and
      Barrault, Lo{\"i}c  and
      Costa-juss{\`a}, Marta R.  and
      Declercq, Christophe  and
      Koponen, Maarit  and
      Kemp, Ellie  and
      Pilos, Spyridon  and
      Forcada, Mikel L.  and
      Scarton, Carolina  and
      Van den Bogaert, Joachim  and
      Daems, Joke  and
      Tezcan, Arda  and
      Vanroy, Bram  and
      Fonteyne, Margot},
    booktitle = "Proceedings of the 23rd Annual Conference of the European Association for Machine Translation",
    month = jun,
    year = "2022",
    address = "Ghent, Belgium",
    publisher = "European Association for Machine Translation",
    url = "https://aclanthology.org/2022.eamt-1.41/",
    pages = "303--304"
}

@article{fleiss1973equivalence,
  title={The equivalence of weighted kappa and the intraclass correlation coefficient as measures of reliability},
  author={Fleiss, Joseph L and Cohen, Jacob},
  journal={Educational and psychological measurement},
  volume={33},
  number={3},
  pages={613--619},
  year={1973},
  publisher={Sage Publications Sage CA: Thousand Oaks, CA}
}

@article{bazrafshan2023role,
  title={The role of ESG ranking in retail and institutional investors' attention and trading behavior},
  author={Bazrafshan, Ebrahim},
  journal={Finance Research Letters},
  volume={58},
  pages={104462},
  year={2023},
  publisher={Elsevier}
}

@inproceedings{koloski-etal-2022-knowledge,
    title = "Knowledge informed sustainability detection from short financial texts",
    author = "Koloski, Boshko and Montariol, Syrielle and Purver, Matthew and Pollak, Senja",
    booktitle = "Proceedings of the Fourth Workshop on Financial Technology and Natural Language Processing (FinNLP)",
    year = "2022",
    address = "Abu Dhabi, United Arab Emirates (Hybrid)",
    publisher = "Association for Computational Linguistics",
}

@article{schimanski2024bridging,
    title = {Bridging the gap in ESG measurement: Using NLP to quantify environmental, social, and governance communication},
    author = {Schimanski, Tobias and others},
    journal = {Finance Research Letters},
    year = {2024},
}

@article{dorfleitner2024esg,
    title = {ESG News Sentiment and Stock Price Reactions: A Comprehensive Investigation via BERT},
    author = {Dorfleitner, Gregor and Zhang, Jun},
    journal = {Schmalenbach Journal of Business Research},
    year = {2024},
}

@article{mehra2022esgbert,
    title = {ESGBERT: Language model to help with classification tasks related to companies environmental, social, and governance practices},
    author = {Mehra, Srishti and Louka, Robert and Zhang, Yixun},
    journal = {arXiv preprint arXiv:2203.16788},
    year = {2022},
}

@article{angioni2024exploring,
    title = {Exploring Environmental, Social, and Governance (ESG) Discourse in News: An AI-Powered Investigation Through Knowledge Graph Analysis},
    author = {Angioni, S. and others},
    journal = {IEEE Access},
    year = {2024},
}

@inproceedings{tseng2023dynamicesg,
    title = {DynamicESG: A Dataset for Dynamically Unearthing ESG Ratings from News Articles},
    author = {Tseng, Yu-Min and Chen, Chung-Chi and Huang, Hen-Hsen and Chen, Hsin-Hsi},
    booktitle = {Proceedings of CIKM},
    year = {2023},
}

@article{lee2024deep,
    title = {Deep-learning-based stock market prediction incorporating ESG sentiment and technical indicators},
    author = {Lee, H. and Kim, J. H. and others},
    journal = {Scientific Reports},
    year = {2024},
}

@misc{Agarwal2025gptoss120bG,
      title={gpt-oss-120b \& gpt-oss-20b Model Card}, 
      author={{OpenAI Team}},
      year={2025},
      eprint={2508.10925},
      archivePrefix={arXiv},
      primaryClass={cs.CL},
      url={https://arxiv.org/abs/2508.10925}, 
}

@inbook{gams, place={Ljubljana}, title={Generative model for less-resourced language with 1 billion parameters}, url={https://repozitorij.uni-lj.si/IzpisGradiva.php?lang=slv&id=164282}, abstractNote={Large language models (LLMs) are a basic infrastructure for modern natural language processing. Many commercial and open-source LLMs exist for English,e.g., ChatGPT, Llama, Falcon, and Mistral. As these models are trained on mostly English texts, their fluency and knowledge of low-resource languages and societies are superficial. We present the development of large generative language models for a less-resourced language. GaMS1B - Generative Model for Slovene with 1 billion parameters was created by continuing pretraining of the existing English OPT model. We developed a new tokenizer adapted to Slovene, Croatian, and English languages and used embedding initialization methods FOCUS and WECHSEL to transfer the embeddings from the English OPTmodel. We evaluate our models on several classification datasets from the Slovene suite of benchmarks and generative sentence simplification task SENTA. We only used af ew-shot in-context learning of our models, which are not yet instruction-tuned. For classification tasks, in this mode, the generative models lag behind the existing Slovene BERT-type models fine-tuned for specific tasks. On a sentence simplification task, the GaMS models achieve comparable or better per formance than the GPT-3.5-Turbo model.}, booktitle={Jezikovne tehnologije in digitalna humanistika}, author={Vreš, Domen and Božič, Martin and Potočnik, Aljaž and Martinčič, Tomaž and Robnik Šikonja, Marko}, year={2024}, pages={485–511}}

@InProceedings{sechidis2011stratification,
author="Sechidis, Konstantinos
and Tsoumakas, Grigorios
and Vlahavas, Ioannis",
editor="Gunopulos, Dimitrios
and Hofmann, Thomas
and Malerba, Donato
and Vazirgiannis, Michalis",
title="On the Stratification of Multi-label Data",
booktitle="Machine Learning and Knowledge Discovery in Databases",
year="2011",
publisher="Springer Berlin Heidelberg",
address="Berlin, Heidelberg",
pages="145--158",
isbn="978-3-642-23808-6"
}

@inproceedings{devlin2019bert,
  title = {{BERT}: Pre-training of Deep Bidirectional Transformers for Language Understanding},
  author = {Devlin, Jacob and Chang, Ming-Wei and Lee, Kenton and Toutanova, Kristina},
  booktitle = {Proceedings of the 2019 Conference of the North American Chapter of the Association for Computational Linguistics (NAACL)},
  pages = {4171--4186},
  year = {2019},
  publisher = {Association for Computational Linguistics},
  url = {https://aclanthology.org/N19-1423/}
}

@inproceedings{liu2019roberta,
  title = {{RoBERTa}: A Robustly Optimized BERT Pretraining Approach},
  author = {Liu, Yinhan and Ott, Myle and Goyal, Naman and Du, Jingfei and Joshi, Mandar and Chen, Danqi and Levy, Omer and Lewis, Mike and Zettlemoyer, Luke and Stoyanov, Veselin},
  booktitle = {Proceedings of the 2019 Conference on Empirical Methods in Natural Language Processing (EMNLP)},
  year = {2019},
  publisher = {Association for Computational Linguistics},
  url = {https://arxiv.org/abs/1907.11692}
}

@inproceedings{conneau2020unsupervised,
  title = {Unsupervised Cross-lingual Representation Learning at Scale},
  author = {Conneau, Alexis and Khandelwal, Kartikay and Goyal, Naman and Chaudhary, Vishrav and Wenzek, Guillaume and Guzmán, Francisco and Grave, Edouard and Ott, Myle and Zettlemoyer, Luke and Stoyanov, Veselin},
  booktitle = {Proceedings of the 58th Annual Meeting of the Association for Computational Linguistics (ACL)},
  pages = {8440--8451},
  year = {2020},
  publisher = {Association for Computational Linguistics},
  url = {https://aclanthology.org/2020.acl-main.747/}
}

@inproceedings{ulcar2021sloberta,
  title = {{SloBERTa}: Slovene Monolingual {BERT}-based Language Model},
  author = {Ul{\v{c}}ar, Matej and Robnik-{\v{S}}ikonja, Marko},
  booktitle = {Text, Speech and Dialogue (TSD 2021)},
  year = {2021},
  publisher = {Springer},
  address = {Cham},
  url = {https://aile3.ijs.si/dunja/SiKDD2021/Papers/Ulcar+Robnik.pdf}
}

@article{kuzman2025teacherstudent,
  title = {{LLM} Teacher-Student Framework for Text Classification with No Manually Annotated Data: A Case Study in {IPTC} News Topic Classification},
  author = {Kuzman, Taja and Ljube{\v{s}}i{\'c}, Nikola},
  journal = {IEEE Access},
  year = {2025},
  publisher = {IEEE},
  url = {https://www.semanticscholar.org/paper/LLM-Teacher-Student-Framework-for-Text-With-No-A-in-Kuzman-Ljube%C5%A1i%C4%87/23436a792cd2b40c69f8f9fd1e112062f66d96ac}
}

@article{nassirtoussi2015text,
  title = {Text mining for market prediction: A systematic review},
  author = {Nassirtoussi, Arman Khadjeh and Aghabozorgi, Saeed and Wah, Teh Ying and Ngo, David C.L.},
  journal = {Expert Systems with Applications},
  volume = {41},
  number = {16},
  pages = {7653--7670},
  year = {2015},
  publisher = {Elsevier},
  url = {https://doi.org/10.1016/j.eswa.2014.06.009}
}

@inproceedings{araci2019finbert,
  title = {{FinBERT}: Financial Sentiment Analysis with Pre-trained Language Models},
  author = {Araci, Dogu},
  booktitle = {Proceedings of the 57th Annual Meeting of the Association for Computational Linguistics: Student Research Workshop},
  pages = {66--71},
  year = {2019},
  publisher = {Association for Computational Linguistics}
}

@article{wang2020minilm,
  title={{MiniLM}: Deep self-attention distillation for task-agnostic compression of pre-trained transformers},
  author={Wang, Wenhui and Wei, Furu and Dong, Li and Bao, Hangbo and Yang, Nan and Zhou, Ming},
  journal={Advances in neural information processing systems},
  volume={33},
  pages={5776--5788},
  year={2020}
}

@misc{team2024gemma,
      title={Gemma 3 Technical Report}, 
      author={{Gemma Team}},
      year={2025},
      eprint={2503.19786},
      archivePrefix={arXiv},
      primaryClass={cs.CL},
      url={https://arxiv.org/abs/2503.19786}, 
}

@article{hollmann2022tabpfn,
  title={{TabPFN}: A transformer that solves small tabular classification problems in a second},
  author={Hollmann, Noah and M{\"u}ller, Samuel and Eggensperger, Katharina and Hutter, Frank},
  journal={arXiv preprint arXiv:2207.01848},
  year={2022}
}

@inproceedings{wolf2020transformers,
  title={Transformers: State-of-the-art natural language processing},
  author={Wolf, Thomas and Debut, Lysandre and Sanh, Victor and Chaumond, Julien and Delangue, Clement and Moi, Anthony and Cistac, Pierric and Rault, Tim and Louf, Remi and Funtowicz, Morgan and others},
  booktitle={Proceedings of the 2020 conference on empirical methods in natural language processing: system demonstrations},
  pages={38--45},
  year={2020}
}

@Article{qian-yang,
AUTHOR = {Qian, Ting and Yang, Caoyuan},
TITLE = {State-Owned Equity Participation and Corporations’ {ESG} Performance in {C}hina: The Mediating Role of Top Management Incentives},
JOURNAL = {Sustainability},
VOLUME = {15},
YEAR = {2023},
NUMBER = {15},
ARTICLE-NUMBER = {11507},
URL = {https://www.mdpi.com/2071-1050/15/15/11507},
ISSN = {2071-1050},
DOI = {10.3390/su151511507}
}


\end{document}